\documentclass[fleqn, 11pt, twoside]{IOEGC}
\usepackage{fourier}

\hypersetup
{
    pdftitle    = {Mechanical Automation with Vision: A Design of Rubik's Cube Solver},
    pdfsubject  = {15th IOE Graduate Conference},
    pdfauthor   = {Abhinav Chalise, Nimesh Gopal Pradhan, Nishan Khanal, Prashant Raj Bista, Dinesh Baniya Kshatri},
    pdfkeywords = {Kociemba, Mechanical system, Rubik's Cube, Stepper Motors, YOLO}
}

\PaperTitle{Mechanical Automation with Vision: A Design for Rubik's Cube Solver\thanks{This is the author’s preprint. The paper was presented at the 15th IOE Graduate Conference, Tribhuvan University, May 2024.}}

\ShortPaperTitle{Mechanical Automation with Vision: A Design for Rubik's Cube Solver}

\Authors{
    Abhinav Chalise \textsuperscript{a}, 
    Nimesh Gopal Pradhan \textsuperscript{b}, 
    Nishan Khanal \textsuperscript{c},
    Prashant Raj Bista \textsuperscript{d}, 
    
    Dinesh Baniya Kshatri \textsuperscript{e}
}

\Abstract{
The core mechanical system is built around three stepper motors for physical manipulation, a microcontroller for hardware control, a camera and YOLO detection model for real-time cube state detection. A significant software component is the development of a user-friendly graphical user interface (GUI) designed in Unity. The initial state after detection from real-time YOLOv8 model (Precision 0.98443, Recall 0.98419, Box Loss 0.42051, Class Loss 0.2611) is virtualized on GUI. To get the solution, the system employs the Kociemba's algorithm while physical manipulation with a single degree of freedom is done by combination of stepper motors' interaction with the cube achieving the average solving time of $\sim$2.2 minutes.
}


\Affiliation{
    \textsuperscript{a} \textsuperscript{b} \textsuperscript{c} \textsuperscript{d} \textsuperscript{e}
    \textit{Department of Electronics and Computer Engineering, Thapathali Campus, IOE, TU}
}

\Email{
    \textsuperscript{a}chalisezabhinav@gmail.com,
    \textsuperscript{b}nimeshgpradhan@gmail.com,
    \textsuperscript{c}khanalnishan28@gmail.com,
    \textsuperscript{d}prashantbista18@gmail.com,
    \textsuperscript{e}dinesh@ioe.edu.np
    }

\begin{document}
    
\maketitle
\thispagestyle{firstpage} 

\section{Introduction}
    The Rubik’s Cube, a 3-D combination puzzle, created in 1974 by Ernő Rubik offers approximately 43 quintillion possible states and presents a significant challenge in pattern recognition and problem-solving although the upper bound to solve the cube was proved to be 20 by Rokicki, et al. \cite{Rokicki_Kociemba_Davidson_Dethridge_2010} fascinating people including researchers, engineers and enthusiasts, known as ‘cubers’, that are actively engaging in competitions, striving to achieve record-breaking times and the fewest move count (FMC) in solving the puzzle.
    
    The challenge of mechanically solving a scrambled Rubik’s Cube has also captured the imagination of innovators and engineers. Various approaches have been attempted, ranging from the development of robotic hands that mimic human movements to the design of specialized cubes that can be rapidly manipulated by machines. These mechanical solutions represent a fascinating intersection of robotics, mechatronics, and computer science, which sets the stage for the development of the ‘Mechanical Automation with Vision: A Design for Rubik’s Cube Solver’, which aims to combine advanced algorithmic processing with a physical mechanism to autonomously solve the Rubik’s Cube.
    
    This research focused on simplicity while maintaining efficiency; reducing the number of motors and overall costs while maintaining the ability to solve the Rubik's Cube within a sensible timeframe. The project's key contributions include:
    \begin{enumerate} 
        \item Development of a Simple Solver that can be easily assembled or created by individuals without needing over-engineered solutions.
        \item The system features an interactive GUI that integrates physical manipulation of a Rubik's Cube and uses advanced object detection for precise component identification.
    \end{enumerate}

\section{Related Works}
    This provides the summary of some of the related works done to tackle the challenge of solving the cube. "The Design of Rubik’s Cube Robot" presents a mechatronics system designed to solve scrambled Rubik’s Cubes including color recognition of each cube face using an industrial CCD camera, followed by the implementation of a two-phase algorithm for cube recovery. Pneumatic manipulators driven by pressurized air, controllable via an electrical relay array, are used to manipulate the cube. \cite{TheDesign_Lu}
    
    "Advanced Rubik’s Cube Algorithmic Solver" utilized a combination of a PC and Arduino Due micro-controller board for processing. Four webcams were strategically placed to capture cube sides, with image processing done via a C\# desktop application using color image segmentation to identify the HSV mask of the cube. Kociemba’s Algorithm, along with the blindfolded method and Old Pochmann, M2, were employed to solve the cube. Six actuators rotated cube faces according to the chosen solution. Webcams transmitted cube side data to the PC, which processed images and generated solutions based on the algorithms. These solutions were then sent to Arduino Due via serial interface, which controlled six stepper motors to rotate cube faces.  \cite{advancedsolver_dan}
    
    "Solving the Rubik’s Cube with Deep Reinforcement Learning and Search" introduces DeepCubeA, a system that merges deep learning with classical reinforcement learning and pathfinding techniques to solve the Rubik’s Cube and similar combinatorial puzzles. DeepCubeA employs approximate value iteration to train a deep neural network, with the goal of approximating the cost-to-reach the goal state. The paper solely focuses on simulating the solving process of the cube and does not involve any hardware implementation. \cite{deepcubea_agostinelli} 
    
    In the paper titled "Rubik’s Cube Solver: A Review,"  a setup comprising six NEMA 17 bipolar stepper motors, L298 motor driver, and Raspberry Pi was utilized. The paper compared various cube-solving algorithms, including Thistlethwaite’s algorithm, Kociemba’s Algorithm, and Korf’s Algorithm. It concluded that Korf’s Algorithm consistently achieved lower move counts and faster speeds compared to other algorithms. \cite{SolverReview_Toshniwal}
    
    A notable gap is seen on hardware simplicity as these works relied on intricate setups, requiring 4 or more motors and multiple webcams leading to increased cost of implementation. Similarly, previous researches lean on conventional cube detection methods without integrating advanced computer vision. Additionally, interactive Graphical User Interface (GUI) where users can interact with the cube and visualize the steps for reaching the solution seems to be absent among them.

\section{System Design and Methodology}
    
    \begin{figure}[H]\centering
    	{\includegraphics[width=0.90\linewidth]{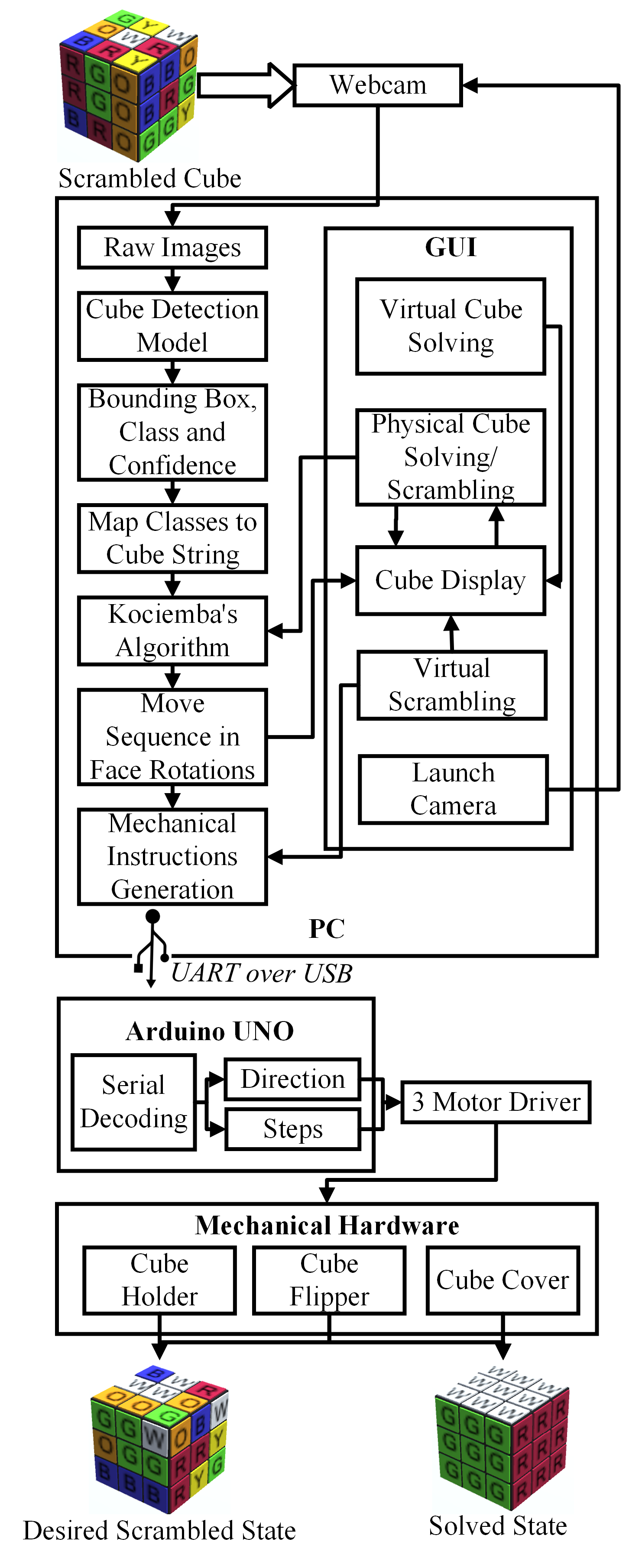}}
    	\caption{System Block Diagram}
    	\label{fig: System Block Diagram}
    \end{figure}
    
    \paragraph{System Block Diagram}
        The system design for solving the Rubik's Cube is outlined in the \textbf{Figure \ref{fig: System Block Diagram}}, which encapsulates the process flows. When the launch camera button is activated in the GUI, the camera captures the cube's scrambled state. This image data is then fed into the YOLO algorithm, which detects individual cubelets and classifies them. These classifications are converted into a string format representative of the cube's state, which is the input for Kociemba's Algorithm. The algorithm processes this data to generate a sequence of fewest face rotations that will lead to a solved cube. This sequence serves a dual purpose: it will map to movements of the mechanical system and updates the virtual cube displayed in the GUI. The scrambles made to virtual cube can also be mapped to the physical cube using Kociemba's algorithm. Communication between the software and the mechanical components is facilitated through Arduino, which has UART serial interface. This orchestrated interaction between the software and hardware components ensures the cube transitions from a scrambled to a solved state or a desired state.
        
        \begin{figure}[H]\centering
        	\fbox{\includegraphics[width=0.90\linewidth]{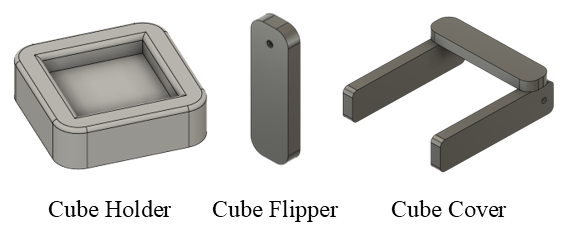}}
        	\caption{3D Parts}
        	\label{fig: 3D Parts}
        \end{figure}
        
\begin{figure}[H]\centering
        	\fbox{\includegraphics[width=0.90\linewidth]{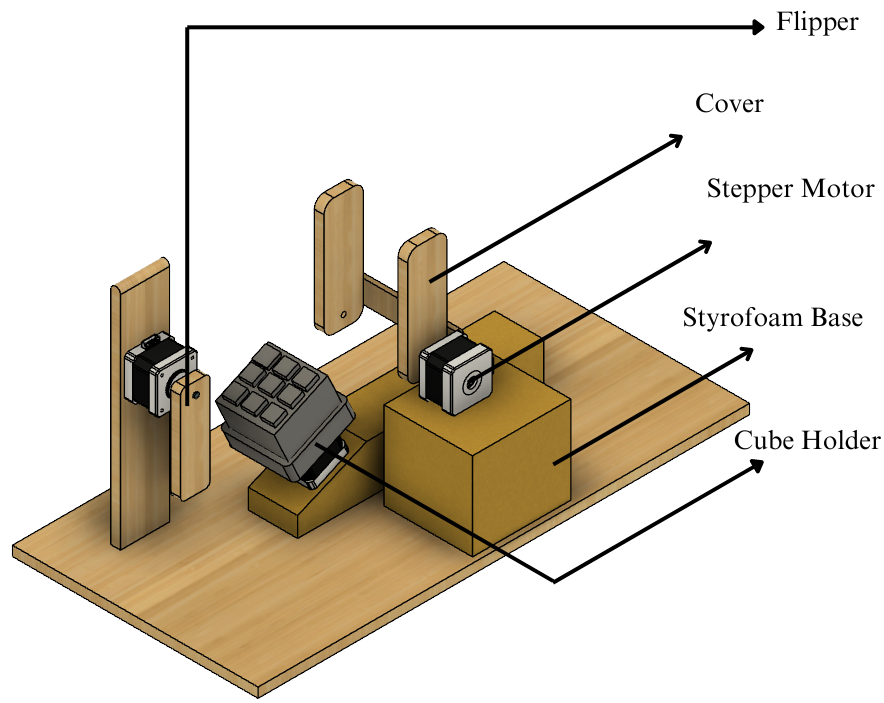}}
        	\caption{Reference 3D Model of System}
        	\label{fig: 3D Model}
        \end{figure}

        \paragraph{Mechanical Design} We constructed a cube cover from wood, with dimensions of $\sim$120x41.2x10 mm, which locks the top two layers of the cube, allowing only the bottom layer to rotate. This cover is connected to a stepper motor shaft and supported by a stand, allowing it to move as needed to either lock the layers or allow full cube rotation. The cube holder holds the Rubik's Cube firmly to prevent any unintentional movement while its inner edges acts as a pivot for the flipping. The holder has an inner square dimension of 60mm and an outer dimension of 70mm, and it's positioned on Styrofoam base tilted at $\sim$10\degree{} for gravity-aided flipping of cube. Lastly, the flipper arm, crafted from wood and measuring $\sim$100x30x10 mm, is vital for reorienting the cube. These carefully crafted components, collaborates in solving the Rubik's Cube. We have also designed 3D printable parts for these three main parts taking our actual parts as reference as shown in \textbf{Figure \ref{fig: 3D Parts}}. The 3D model in \textbf{Figure \ref{fig: 3D Model}} is a reference for actual design which will aid future recreations.
        
\begin{figure*}[hbt]\centering
            \fbox {\includegraphics[width=0.92\linewidth]{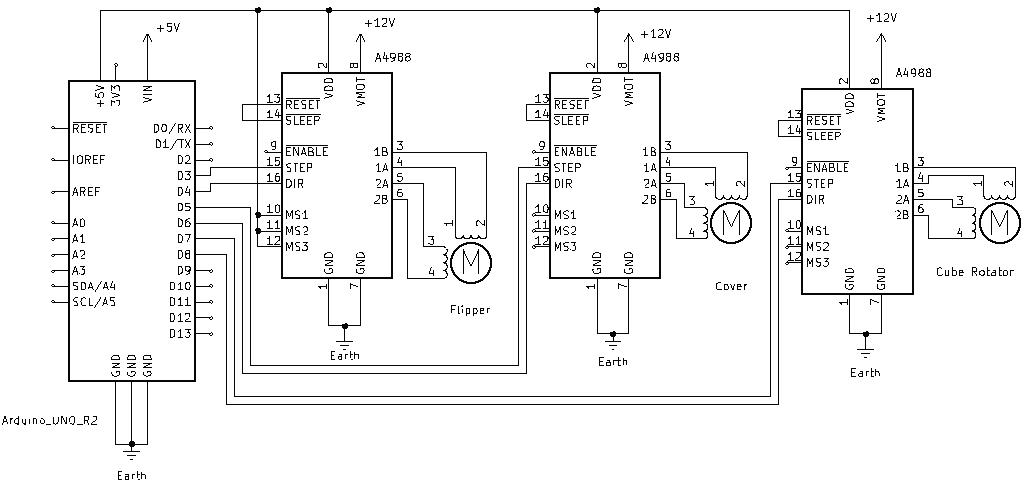}}
            \caption{Circuit Diagram}
            \label{fig: Circuit}
        \end{figure*}

        \paragraph{Circuit Design} The circuit design in \textbf{Figure \ref{fig: Circuit}} illustrates an Arduino UNO connected to three A4988 stepper motor driver modules to control three stepper motors. The Arduino, powered by a +5V supply, interfaces with each A4988 module through directional (DIR) and step (STEP) pins to control motor direction and step sequences, respectively. Additionally, micro-stepping resolution is set via MS1, MS2, MS3 pins. Both the +5V and ground (GND) lines from the Arduino connect to the VDD and ground pins on the A4988 modules to power the logic circuitry. Each A4988 receives a +12V supply to power the stepper motors and has its ground connected to a common ground. Outputs from the drivers (2B, 2A, 1A, 1B) connect to the respective stepper motor coils, initiating controlled movements of the Flipper, Cover, and Cube Holder motors as per the signals from the Arduino.
        
        \paragraph{Cube Solving Algorithm} To facilitate the solving of a Rubik's cube, Kociemba's algorithm is utilized, implemented in Python. The algorithm is initialized with a maximum depth (d) of 24 moves, which is the maximum steps to solution for the given implementation. This initialization ensures that the algorithm searches for a solution within this optimal move count as shown by the following pseudocode. \cite{rokicki2008twentyfive}
        
        \textbf{Kociemba's Algorithm: Pseudocode} 
        
        Require: Max depth \textbf{d}, Bound \textbf{b}, Current cube state \textbf{s}\\
        $1:d \leftarrow 0$\\
        $2:b \leftarrow \infty$\\
        $3:$ \textit{while} $d<b$ \textit{do}\\
        $4:$ \ \ \textit{for} $s \in S^{d},\ r(ps) = e$ \textit{do}\\
        $5:$ \ \ \ \textit{if} $d + d_{2}(ps) < b$ \textit{then}\\
        $6:$ \ \ \ \ \textit{Solve phase} $2$\textit{; report new better solution}\\
        $7: \ \ \ \ b = d + d_{2}(ps)$\\
        $8:$ \ \ \textit{end if}\\
        $9:$ \ \textit{end for}\\
        $10: d \leftarrow d + 1$\\
        
\begin{figure*}[hbt]\centering
        	\fbox {\includegraphics[width=0.90\linewidth]{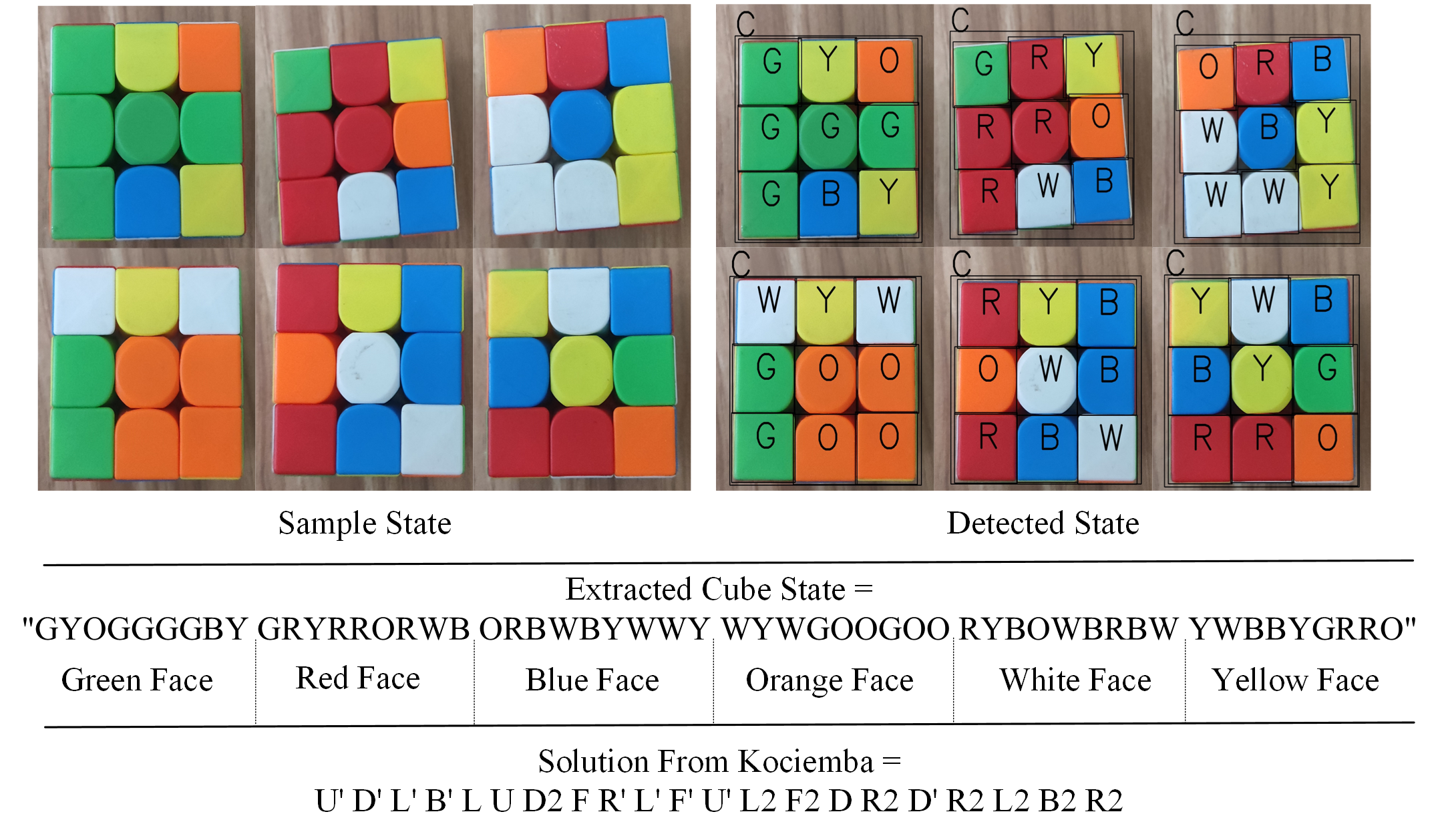}}
        	\caption{Detection and Solution of a Sample Cube}
        	\label{fig: SampleDetection}
        \end{figure*}

        $\textbf{S}^\textbf{d} =$ \textit{set of all states at depth d}\\
        $\textbf{ps} =$ \textit{current state of cube}\\
        $\textbf{r(ps)} =$ \textit{function to determine whether state ps should be excluded from further exploration or not}\\
        $\textbf{d}_\textbf{2}\textbf{(ps)} =$ \textit{function that estimates distance from state ps to solved state}
        \paragraph{Cube State Detection} We selected a YOLOv8 model which was trained on COCO dataset. In the final training phase, we fine-tuned the model by adjusting specific hyperparameters, including setting epochs to 70, learning rate to 0.000909, image size to 1120, utilizing the yolov8n.pt model with an AdamW optimizer and a momentum of 0.9. A custom YOLO cube detection model has been developed, which integrates with OpenCV for real-time image processing. When fed an image, the model swiftly identifies cubes and cubelets, predicting their class and bounding box. These predictions are then used to derive the cube's state string for each face. Upon processing all six faces, the individual face state strings are combined to form the complete cube state string, illustrated in \textbf{Figure \ref{fig: SampleDetection}}, which is handed over to Kociemba's algorithm to generate the solution string.
        
\begin{figure*}[hbt]\centering
        	\fbox {\includegraphics[width=0.86\linewidth]{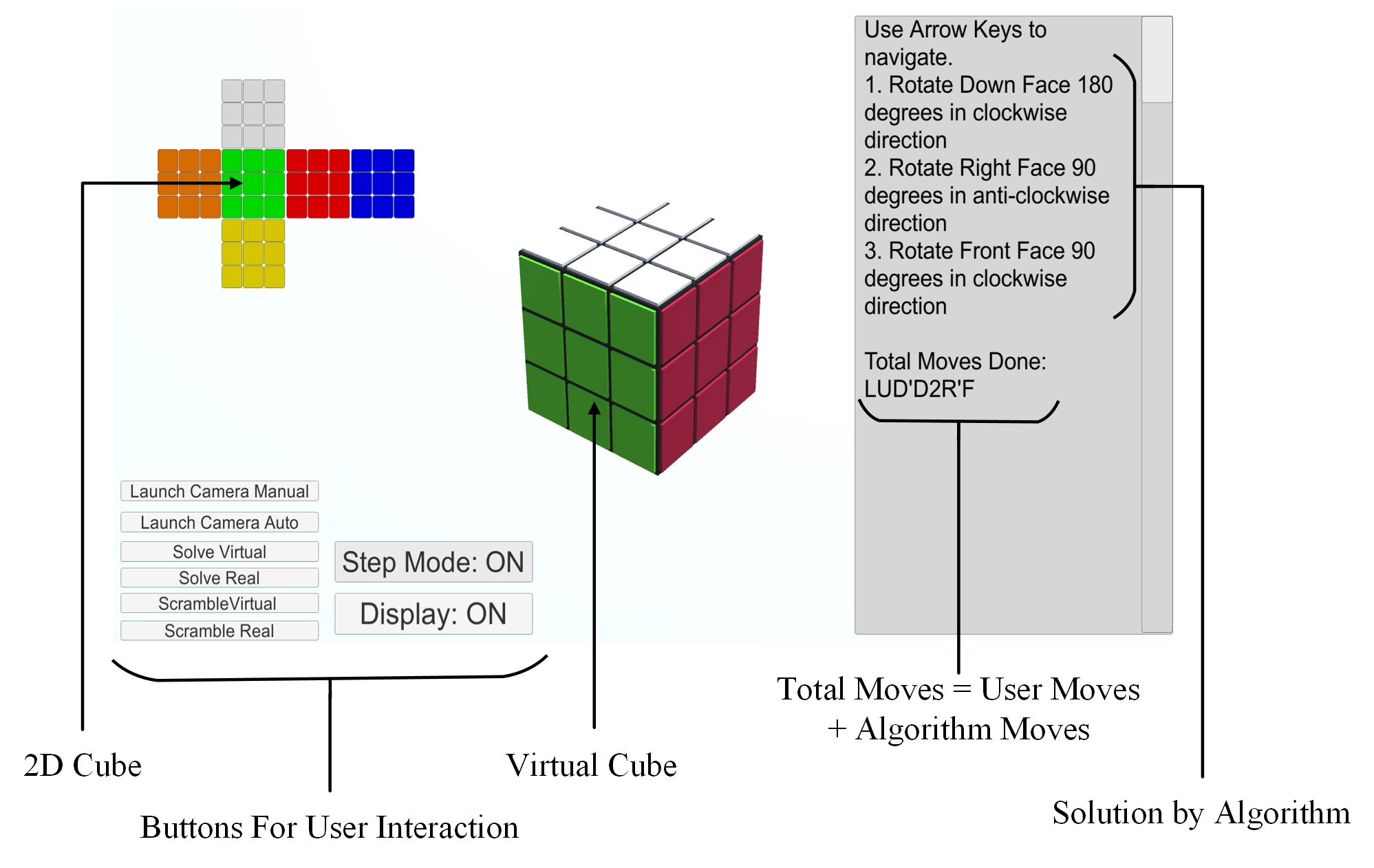}}
        	\caption{Graphical User Interface}
        	\label{fig: GUI_2}
        \end{figure*}

        \paragraph{GUI} The GUI software was designed in Unity which includes a UI to interact with the cube and choose the solving process. Users can toggle the 'Display' option to show or hide the solution panel. The solution panel shows the moves generated by the algorithm and the total moves. The total moves comprises of the user's attempted moves to solve the cube and the algorithm generated moves following the user's attempt. A right hand clockwise face rotation is represented by capital letter of the face name while an apostrophe denotes counter-clockwise rotation. The number 2 after the letter represents two rotations in the same direction. For example, in \textbf{Figure \ref{fig: GUI_2}}, LUD' represents the user moves i.e. 90\degree{} clockwise rotation of left and upper face followed by 90\degree{} anticlockwise rotation of down face and D2R'F is the algorithm generated moves which is elaborated in the panel. The ‘Launch Camera Auto' button automatically captures and displays the state of cube using a webcam. A ‘Solve Virtual' button applies Kociemba's algorithm to a virtual cube, which users can scramble using their mouse or keyboard. The ‘Launch Camera Manual' button also captures the cube's state, but requires the user confirmation for each face. The ‘Solve Real' button extends the functionality of Kociemba's algorithm to both the virtual and the real cube based on the captured state. There are also ‘Scramble Virtual' and ‘Scramble Real' buttons for randomizing the state of the virtual and/or real cubes, respectively. Lastly, a ‘Step Mode' allows users to step through the solving process with arrow keys, giving them control over the pace of solving.
    
    \subsection{Dataset Preparation}
\begin{figure}[H]\centering
                \fbox{\includegraphics[width=0.92\linewidth]{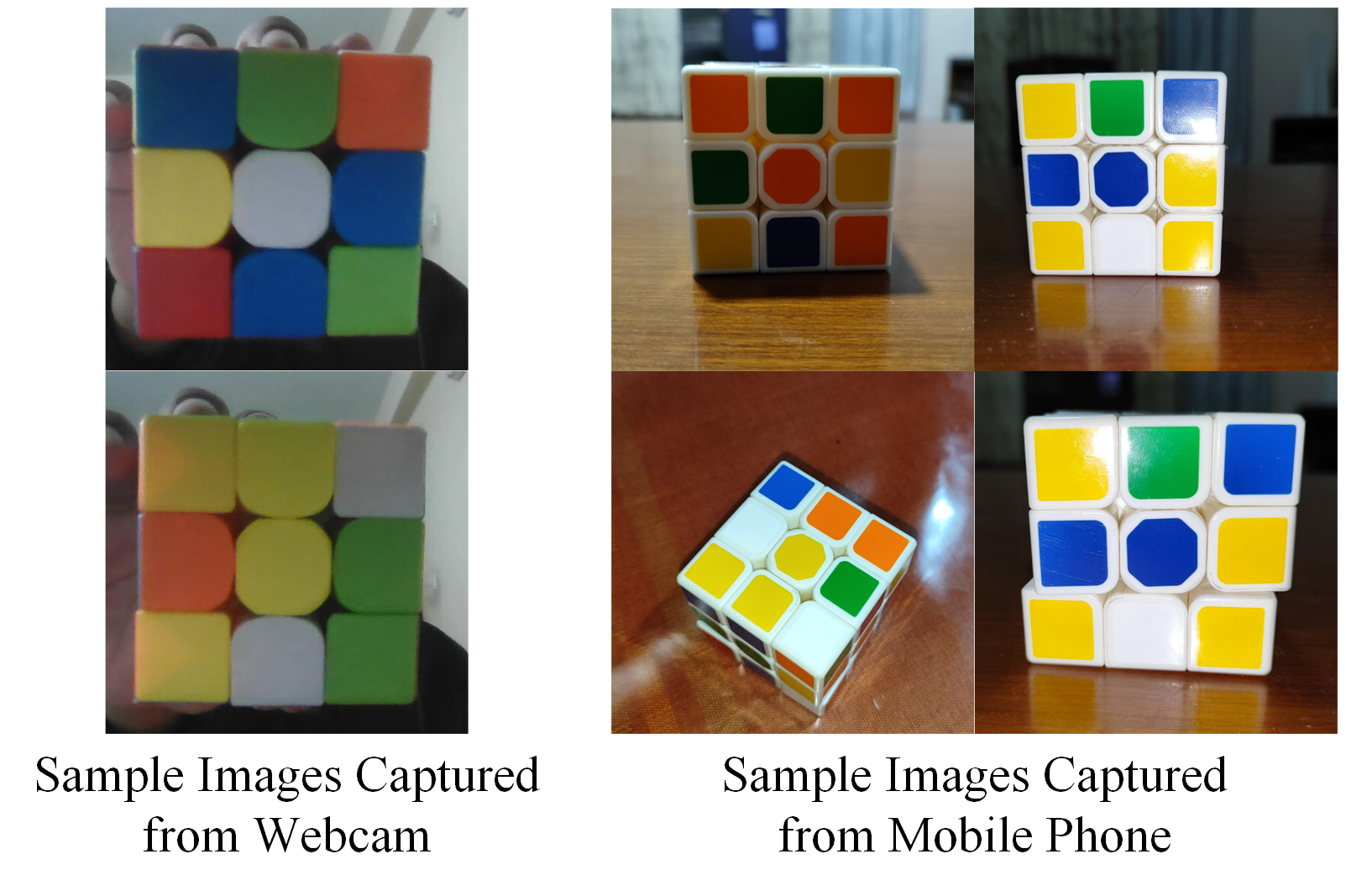}}
                \caption{Sample Dataset Images}
                \label{fig: Sample Images}
            \end{figure}

        \paragraph{Dataset Collection and Augmentation}
             The dataset for training the cube detection algorithm was meticulously compiled by capturing images of cubes through a laptop webcam and Samsung A52 \& Poco X5 phone as seen on \textbf{Figure \ref{fig: Sample Images}}. A total of 350 unique images were obtained during this manual collection process. Python script was employed to systematically capture images from the webcam and appropriately name them. Additionally, for images captured with the phone, ffmpeg program was employed to compress the larger-sized images.

             To enhance the diversity of the dataset, augmentation techniques were applied to the collected images and a Python script was utilized to alter brightness, saturation, exposure, and contrast, (altering $\alpha$ and $\beta$ values and HSV values of image) resulting in $\sim$900 total images that enrich the dataset.
             
            \begin{figure}[H]\centering
            	\fbox{\includegraphics[width=0.88\linewidth]{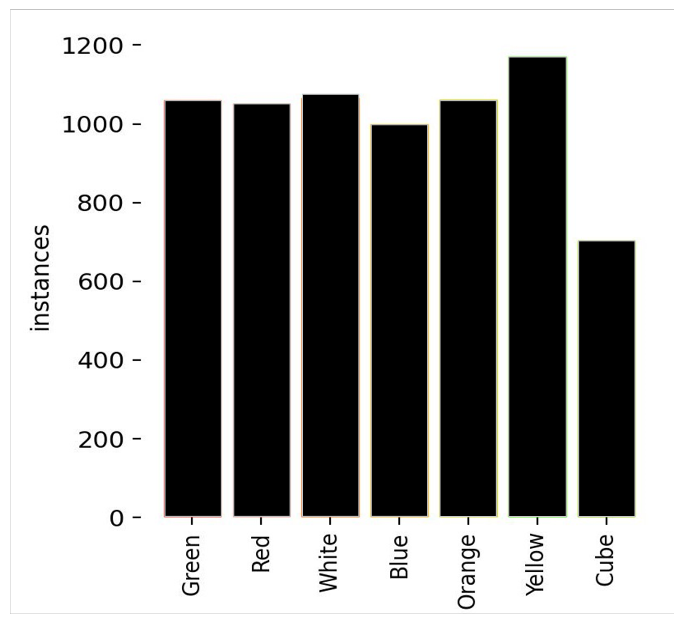}}
            	\caption{Histogram of Training Set}
            	\label{fig: Histogram}
            \end{figure}
            
        \paragraph{Image Annotation and Organization}
             For effective training of the cube detection algorithm, images were annotated to label the distinct objects within them. Seven distinct classes were defined (cube and six colors). The annotation process was carried out using the open-source annotation tool, CVAT.

             To train the model, the dataset was split into training and validation sets in an 80:20 ratio. Approximately \textbf{700 images} were allocated for \textbf{training}, while \textbf{200 images} were reserved for \textbf{validation}. The histogram in \textbf{Figure \ref{fig: Histogram}} shows the instances of different classes for training dataset.

\section{Results and Discussion}
    \subsection{Detection Model}
        The results associated with detection model training include loss metrics, evaluation metrics and confusion matrix. From \textbf{Table \ref{tbl:YOLO Losses}}, a rapid and steep decline in losses from epoch 1 to epoch 30 can be seen, then the rate decreases and starts to saturate. This shows that the model was trained effectively and it's learning to predict bounding boxes and its corresponding classes accurately. 

        \textbf{Table \ref{tbl: YOLO Performance}} shows that Precision, Recall, and mAP50 metrics improve rapidly over the first few epochs, indicating initial learning progress. After around the 10th epoch, all metrics saturate, suggesting that the model has mostly converged and the model is able to perform accurately and consistently.

\begin{figure}[H]\centering
        	\fbox{\includegraphics[width=0.92\linewidth]{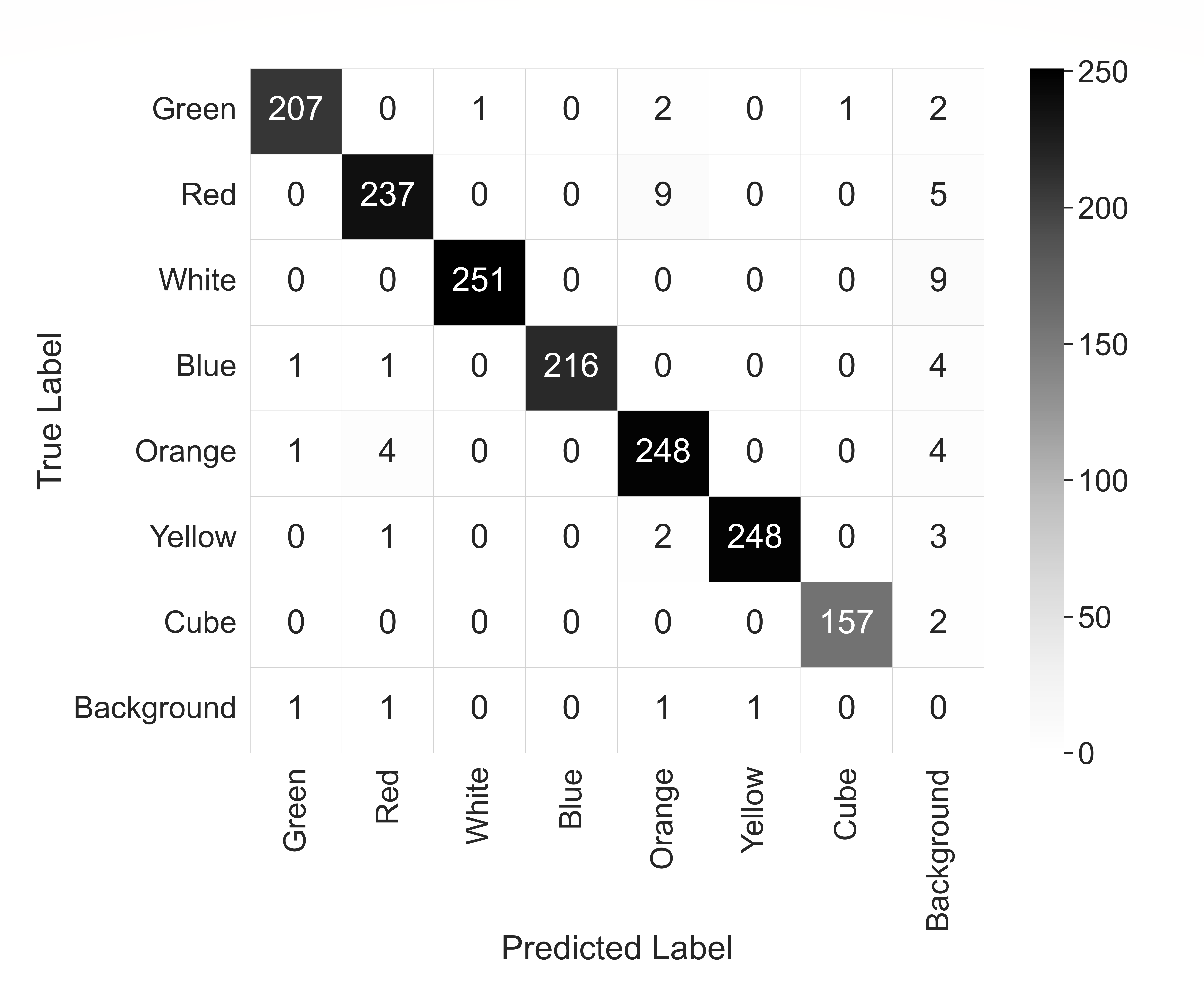}}
        	\caption{Confusion Matrix}
        	\label{fig: Confusion Matrix}
        \end{figure}

        \textbf{Figure \ref{fig: Confusion Matrix}} shows that the classification model performs well, with high accuracy for most classes as tested with validation dataset of $\sim$200 images. Some confusion exists between "Red" and "Orange" due to their color similarity. Lighting condition also affects the detection of cube, with drastic conditions rendering detection impossible.
    
        \begin{table}[H]
        	\caption{YOLOv8 Loss Metrics}
        	\label{tbl:YOLO Losses}
        	\centering
        	\begin{tabular}{|c|c|c|c|c|} 
        		\hline
        		\textbf{Epoch} & \textbf{TBL} & \textbf{TCL} & \textbf{VBL} & \textbf{VCL} \\
        		\hline
        		1 & 0.95719 & 2.9183 & 0.88011 & 3.0461\\
        		5 & 0.79959 & 0.89576 & 0.83025	& 0.82605\\
        		10 & 0.72736 & 0.6779 & 0.75193 & 0.61718\\
        		15 & 0.64833 & 0.56015 & 0.66463 & 0.49791\\
        		20 & 0.60223 & 0.48648 & 0.65736 & 0.4541\\
                25 & 0.56207 & 0.42682 & 0.56694 & 0.39193\\
                30 & 0.52661 & 0.40773 & 0.55289 & 0.36016 \\
                35 & 0.48895 & 0.37072 & 0.53207 & 0.34346\\
                40 & 0.4668 & 0.3382 & 0.51037 & 0.33165\\
                45 & 0.4425 & 0.31981 & 0.49005 & 0.3139\\
                50 & 0.42816 & 0.30919 & 0.46641 & 0.2888\\
                55 & 0.39984 & 0.28916 & 0.46243 & 0.28574\\
                60 & 0.39997 & 0.28485 & 0.44754 & 0.2776\\
                65 & 0.34868 & 0.24467 & 0.43081 & 0.27128 \\ 
                70 & 0.32772 & 0.23121 & 0.42051 & 0.2611\\
        		\hline
        	\end{tabular}
        \end{table}
        \textit{(T/V)BL} $=$ \textit{(Train/Validation) Box Loss}\\
        \textit{(T/V)CL} $=$ \textit{(Train/Validation) Class Loss}
    
        \begin{table}[H]
            \caption{YOLOv8 Performance Evaluation Metrics}
            \label{tbl: YOLO Performance}
            \centering
            \begin{tabular}{|c|c|c|c|}
                \hline
                \textbf{Epoch} & \textbf{Precision} & \textbf{Recall} & \textbf{mAP50} \\
                \hline
                1   & 0.17353 & 0.6606  & 0.29833 \\
                5   & 0.8963  & 0.93123 & 0.94455 \\
                10  & 0.9182  & 0.96391 & 0.96295 \\
                15  & 0.94434 & 0.96763 & 0.97933 \\
                20  & 0.97034 & 0.98027 & 0.98385 \\
                25  & 0.97118 & 0.9751  & 0.98898 \\
                30  & 0.97767 & 0.97192 & 0.98921 \\
                35  & 0.97076 & 0.98069 & 0.99022 \\
                40  & 0.97619 & 0.97561 & 0.98936 \\
                45  & 0.97298 & 0.98215 & 0.98773 \\
                50  & 0.98063 & 0.98694 & 0.99049 \\
                55  & 0.98438 & 0.97878 & 0.98984 \\
                60  & 0.98259 & 0.97835 & 0.99119 \\
                65  & 0.98213 & 0.9842  & 0.98954 \\
                70  & 0.98443 & 0.98419 & 0.99047 \\
                \hline
            \end{tabular}
        \end{table}

    \subsection{Mechanical System and Live Detection}
        
\begin{figure}[H]\centering
        	\fbox{\includegraphics[width=0.92\linewidth]{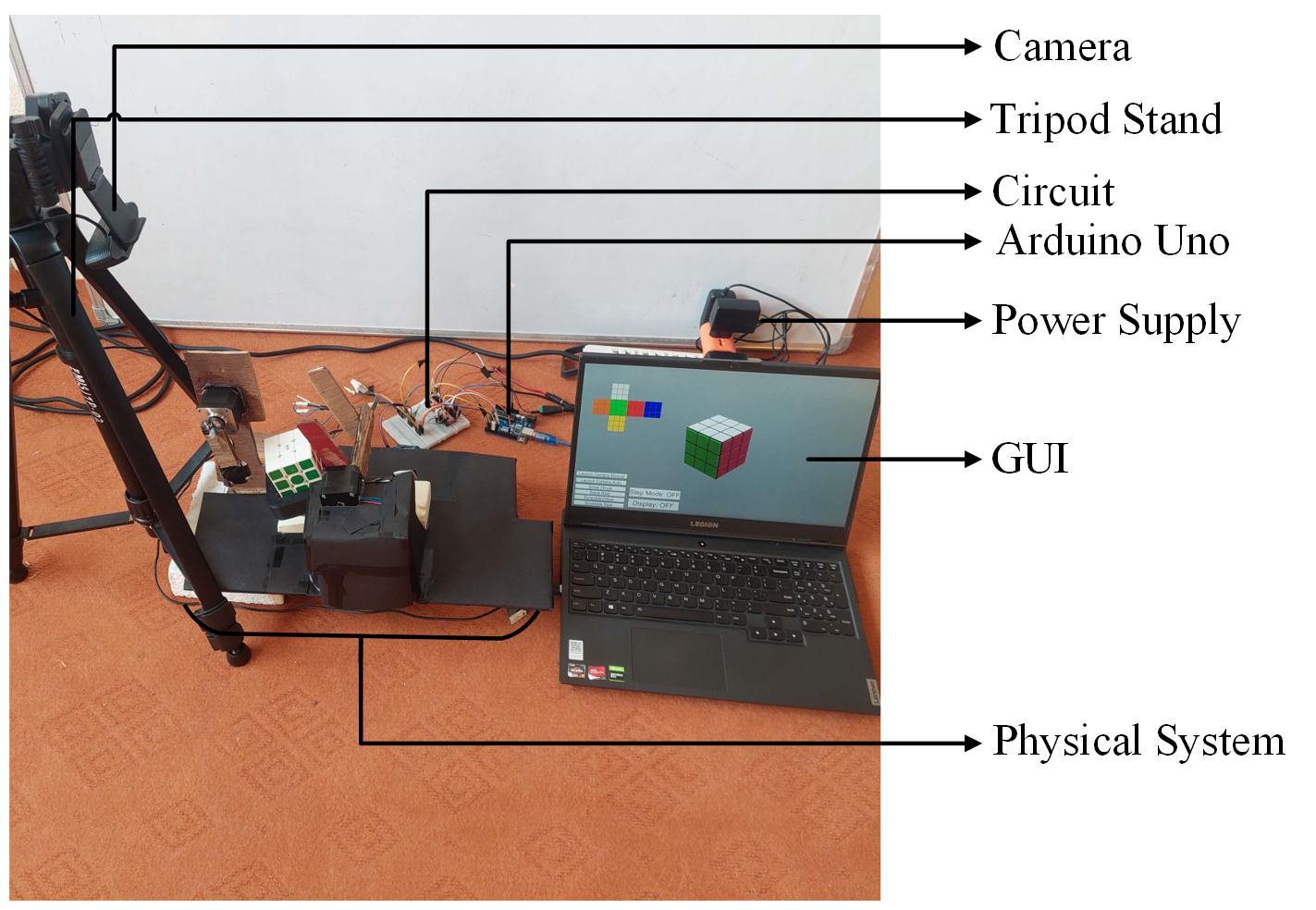}}
        	\caption{Rubik's Cube Solver}
        	\label{fig: Solver1}
        \end{figure}

        The resulting solver is shown in \textbf{Figure \ref{fig: Solver1}} which shows the overall system integration between mechanical system, camera, Arduino and computer. A working demo for the system is available at: \href{https://youtu.be/rlcDXjqy2Vs}{https://youtu.be/rlcDXjqy2Vs}

        The time taken for execution of each move by the mechanical system was set fixed to balance the consistency and speed which is shown in \textbf{Table \ref{tbl:Time to Complete Individual Moves}}.
                
        The system's performance was evaluated through 1000 solution moves, each ranging from 18 to 24 face rotations in length. Then the mechanical moves for realizing the solutions were generated through a combination of moves mentioned in Table \ref{tbl:Time to Complete Individual Moves} and the average time taken to execute these moves was 128366.108ms($\sim$2.2 minutes). \textbf{Table \ref{tbl: Time To Execute Solution Moves}} presents some of the sample data taken from the total of 1000. 
        
        \begin{table}[H]
            \caption{Time to Complete Individual Moves}
            \label{tbl:Time to Complete Individual Moves}
            \centering
            \begin{tabular}{|p{0.7\linewidth}|p{0.2\linewidth}|}
                \hline
                \textbf{Name of the Move} & \textbf{Time (ms)} \\
                \hline
                Flipping & 2731 \\
                \hline
                Rotating whole cube 90 degrees & 1074 \\
                \hline
                Rotating bottom layer 90 degrees clockwise direction & 2028 \\
                \hline
                Rotating bottom layer 90 degrees anticlockwise direction & 2582 \\
                \hline
                Rotating bottom layer 180 degrees (s) & 3319 \\
                \hline
            \end{tabular}
        \end{table}
    
    \begin{table}[H]
        \caption{Time to Execute Solution Moves }
        \label{tbl: Time To Execute Solution Moves}
            \centering
        \begin{tabular}{|p{0.7\linewidth}|p{0.2\linewidth}|}
            \hline
            \textbf{Face Rotation Solution Moves} & \textbf{Time (ms)} \\
            \hline
            R' B U2 B D L F2 R' B' D' U' U' F' F U L' D2 F2 B2 R B U L2 & 140967 \\
            \hline
            U B' D F U2 R B' L' F2 R2 D' D2 B2 B2 F2 L D2 L' F2 L' L U2 & 132163 \\
            \hline
            L2 R' L D' F2 D2 F F' U2 R' U D2 D' U R2 D D2 L' F F D' B & 134797 \\
            \hline
            U F2 L' D2 D2 R2 B L' R' D D2 B' D D R' U2 F' R2 B' B2 & 115738 \\
            \hline
            L2 L2 B D2 U R B2 D R U U' D' D2 R F2 L' B' R' B' U2 U2 L2 L2 & 131460 \\
            \hline
        \end{tabular}
    \end{table}

     \begin{figure*}[hbt]\centering
    	\fbox {\includegraphics[width=0.90\linewidth]{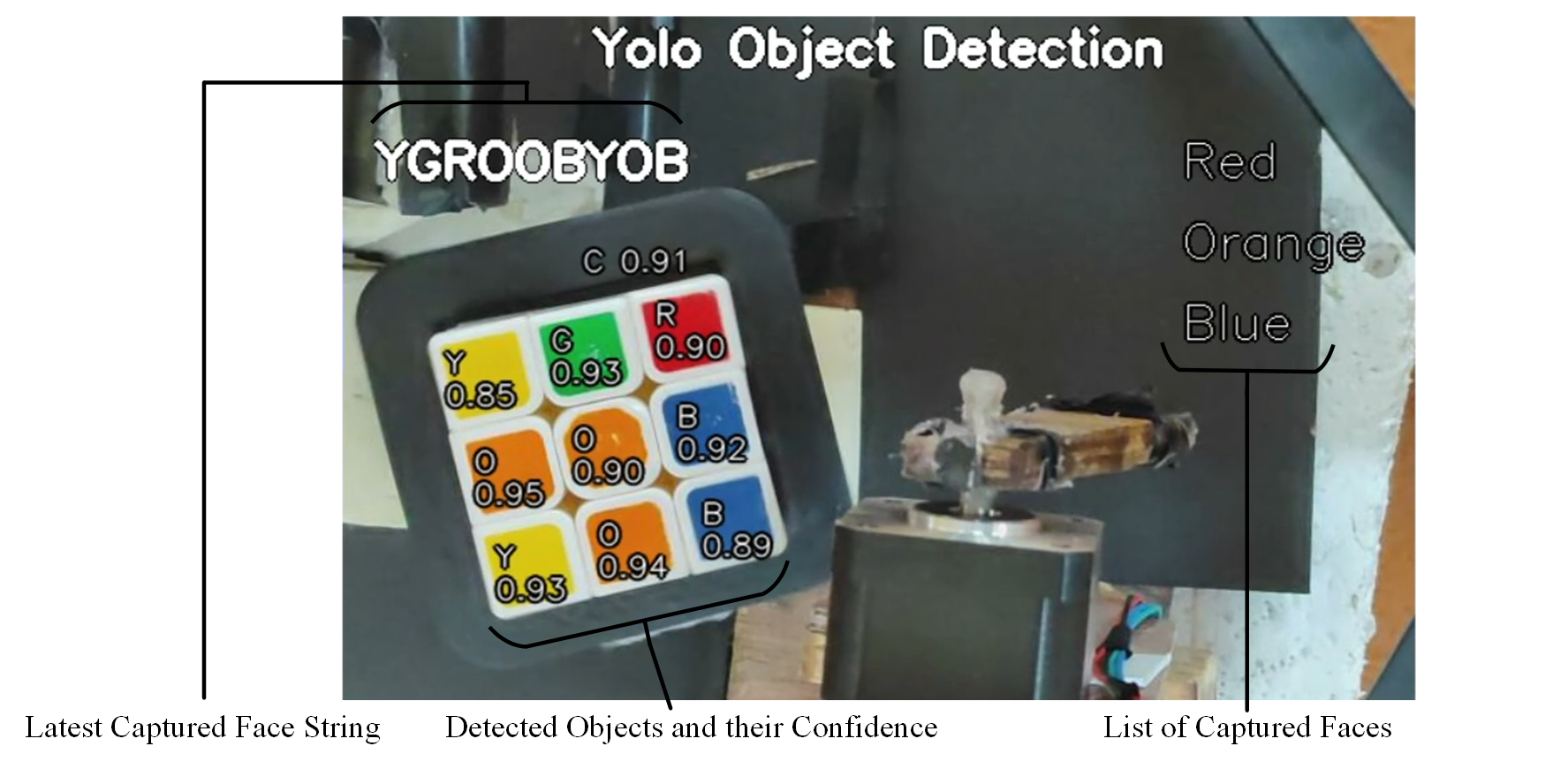}}
    	\caption{Live Cube Detection}
    	\label{fig: Live Detection}
    \end{figure*}
    
    \paragraph{Live Detection}
     In \textbf{Figure \ref{fig: Live Detection}}, the YOLO object detection program shows the detected objects alongside their confidence scores. In the top-left corner, the program shows the face string of the latest captured cube face, while the right side presents a list of faces already captured.
    
    \subsection{Calibration of Mechanical System}
     The functionality of our cube-solving model depends on its initial starting state. Utilizing stepper motors which are an open-loop system, the model’s angles of rotation remain fixed once set, regardless of subsequent states. Improper starting position will cause cube to fall off the holder. The calibration procedures ensure the accurate and consistent operation of our cube-solving model.
    
     To calibrate the cover, initially it is positioned parallel to the motor, establishing our baseline. From this baseline point, the cover undergoes required rotation to its starting point.
    
     For calibrating the cube holder markings has been made on the styrofoam base which serves as a guideline. Initially, the cube holder is aligned with the markings, establishing our baseline position then the cube holder undergoes required rotation to its starting point.
     
\section{Conclusion}
This study introduces a vision-aided autonomous Rubik's Cube solver designed with a minimalist hardware configuration and user-friendly graphical interface (GUI). The research presents an integration of three disciplines; machine learning, software development, and hardware implementation, contributing the advancement of robotics and machine learning by providing insights into hardware optimization, software development and 3D design. Additionally, the system's efficiency could be enhanced through the implementation of a closed-loop feedback mechanism, leveraging magnetic auto-encoder technology.

\phantomsection
\section*{Acknowledgements} 
\addcontentsline{toc}{section}{Acknowledgements} 
The authors extend their gratitude to the Department of Electronics and Computer Engineering, Thapathali Campus, and the Robotics and Automation Center, Thapathali for their invaluable support and resources, which were instrumental in the completion of this research.

\phantomsection
\bibliographystyle{unsrt}
\bibliography{refs}

\end{document}